\documentclass{article}

\usepackage{arxiv}

\usepackage[utf8]{inputenc} 
\usepackage[T1]{fontenc}    
\usepackage{hyperref}       
\usepackage{url}            
\usepackage{booktabs}       
\usepackage{amsfonts}       
\usepackage{nicefrac}       
\usepackage{microtype}      
\usepackage{lipsum}

\usepackage{subcaption}
\usepackage{rotating}
\usepackage{tikz}
\usepackage{lscape}
\usetikzlibrary{positioning}
\usepackage{array,calc}
\newlength\lena \newlength\lenb \newlength\lenc \newlength\lend
\newcolumntype{P}[1]{>{\centering\arraybackslash}p{#1}} 
\newcommand\mcii[1]{\multicolumn{8}{P{\lend}|}{#1}}  
\newcommand\mciii[1]{\multicolumn{16}{P{\lenc}|}{#1}}

\title{A semi-supervised deep residual network for mode detection in Wi-Fi signals}

\author{
  Arash Kalatian \\
  Laboratory of Innovations in Transportation (LiTrans)\\
  Ryerson University\\
  Toronto, Canada \\
  \texttt{arash.kalatian@ryerson.ca} \\
   \And
  Bilal Farooq \\
  Laboratory of Innovations in Transportation (LiTrans)\\
  Ryerson University\\
  Toronto, Canada \\
  \texttt{bilal.farooq@ryerson.ca} \\
}

\begin{document}
\maketitle

\begin{abstract}
 Due to their ubiquitous and pervasive nature, Wi-Fi networks have the potential to collect large-scale, low-cost, and disaggregate data on multimodal transportation. In this study, we develop a semi-supervised deep residual network (ResNet) framework to utilize Wi-Fi communications obtained from smartphones for the purpose of transportation mode detection. This framework is evaluated on data collected by Wi-Fi sensors located in a congested urban area in downtown Toronto. To tackle the intrinsic difficulties and costs associated with labelled data collection, we utilize ample amount of easily collected low-cost unlabelled data by implementing the semi-supervised part of the framework. By incorporating a ResNet architecture as the core of the framework, we take advantage of the high-level features not considered in the traditional machine learning frameworks. The proposed framework shows a promising performance on the collected data, with a prediction accuracy of 81.8\% for walking, 82.5\% for biking and 86.0\% for the driving mode.
\end{abstract}

\keywords{Mode detection, Wi-Fi signals, semi-supervised learning, ResNet, Deep Neural Networks}



\section{Introduction}
\label{S:1}
Conventionally, self-reported surveys have been the main source for collecting transportation data from network users. Although these methods have been employed for decades, their intrinsic problems, as well as the recent advances in location-aware technologies, have made researchers rethink conventional travel survey techniques \cite{chen2010evaluating}. Some of the main problems of traditional surveys include: being time-consuming, expensive and not representative, and involvement of human error and biased responses \cite{murakami2004using,gong2012gps,stopher2007household}.

Location-aware technologies and networks, on the other hand, can potentially be used for ubiquitous data collection at large scales and in different conditions. High penetration rate of smartphones allows collecting data from an ample number of users, who may not even be aware of the experiment. For instance, by using Wi-Fi sensors, data can be collected from all the participants carrying a Wi-Fi enabled device in the area of the experiment \cite{farooq2015ubiquitous}. As Wi-Fi and Bluetooth sensors are already operational in some urban areas, e.g. city of Toronto \cite{opendata}, no additional infrastructural costs are required for such studies.

In transportation studies, mode detection is of interest as it helps city planners and transportation agencies to observe and track shares of different transportation modes over time. This information can then be exploited for planning, designing, and operating multi-modal infrastructures required by traffic network users. Information derived based on modes can also be utilized in other fields, such as contextual advertisements, health applications (e.g. steps and calorie counters) and environmental studies (e.g. carbon footprints).

Supervised learning algorithms have been the dominant tools to infer mode of transportation from the collected data in the literature. In such case, labelled records for each trip, i.e. mode of transportation for each trip, are required to train and validate algorithms. Thus, data should either be manually labelled by looking into video footage, for instance, or be limited to selected participants who would label their own trips. In both cases, the advantage of large scale data collection will be negatively affected as the number of data points obtained will be significantly reduced. To benefit from large amounts of unlabelled data available, semi-supervised or unsupervised algorithms can be taken into account. In basic terms, semi-supervised learning algorithms couple a small-sized labelled data with unlabelled data for the purpose of training a classifier~\cite{blum1998combining}. 

In this study, we develop a semi-supervised deep residual neural network for Wi-Fi signals to infer mode of transportation of network users in a congested urban area in Downtown Toronto. In recent years, Deep Neural Networks (DNN) have demonstrated successful performances in different fields of machine learning. DNNs, if coupled with adequate procedures, have shown impressive performances on complex and noisy data \cite{reed2014training,rolnick2017deep}. However, increasing the number of hidden layers after some point results in the problem of degradation in accuracy \cite{he2016deep}. This concern led to the introduction of impressively successful Deep Residual Networks (ResNet), which tries to increase the number of hidden layers by implementing \textit{Shortcut Connections} \cite{he2016deep}.

The rest of this paper is organized as follows: In the next section, we review previous researches on the subject. Section \ref{S:4} presents data collection and pre-processing procedures in detail. We then describe the framework and architecture of our proposed algorithm in section \ref{S:3}. We apply our proposed framework on the data collected in section \ref{S:5}. In the end, conclusions and future research plans are outlined in section \ref{S:6}.

\section{Literature Review}
\label{S:2}
 Data collection methods that are based on location-aware technologies, and thus the associated mode detection studies, can be divided into two main categories: user-centric and network-centric methods. In \emph{user-centric} methods, users are required to be actively involved in the data collection procedure. Examples include GPS and accelerometer data, or a combination of both that can be used to infer transportation mode \cite{zheng2008understanding,reddy2008determining,stenneth2011transportation,endo2016deep,xiao2017identifying,dabiri2018inferring}. 
\cite{zheng2008understanding} used GPS data to detect mode of transportation. In their study, features such as heading change rate, stop rate and velocity change rate were defined. A graph-based post-processing algorithm was proposed considering the conditional probability between different modes of transport. Using the approach suggested in this study, transportation mode of users in 76\% of the experiments was predicted correctly. Using multiple data sources has also been widely practiced in related studies. \cite{reddy2008determining} implemented GPS data along with smartphones accelerometer data to identify walking, running, biking and motorized modes. An accuracy of over 90\% was achieved using two-stage decision tree and discrete Hidden Markov Model. The data collection procedure of this study involved attaching 5 phones to each participant simultaneously, which seems to be impossible to implement for real-life large scale-data collection. Moreover, training classifiers based on data from 5 out of 6 participants and validating them on data from the other participant, resulted in a drop in accuracy by 10\%, which may have been due to the overfitting to training data. \cite{stenneth2011transportation} added transportation network information to determine users' mode of transport between stationary, walking, biking, driving, and using public transit. Classifiers such as decision tree, random forest, and Multilayer Perceptron were developed in this study. Training classifiers using solely GPS data resulted in an accuracy of 75\%. However, when network information such as bus routes, rails, real-time bus schedules, etc, were added, random forest method was shown to have an accuracy of 93.7\%. Using multiple data sources may not always be an available option in different cities. Thus making this approach difficult to introduce as a generalized solution. \cite{xiao2017identifying} outperformed traditional decision tree-based models by applying a tree-based ensemble classification algorithms using over 100 GPS trajectory augmented features .
Although user-centric approaches potentially tend to collect more precise and accurate data, the collection relies upon a specific and limited number of participants. This makes such methods hard to implement in large scales and to address real-life transportation problems. Such data collection method may result in biased data due to the involvement of only a certain type of participants. In addition, extra operational costs are usually associated with such studies, as they require mobile Apps, participant's time, and a high level of battery consumption. 

On the contrary, \emph{network-centric} methods try to collect data passively, requiring no intervention from the users of the network. Main sources of data in network-centric approaches in the literature have been Wi-Fi data, Bluetooth transceivers data, and GSM signals data~\cite{sohn2006mobility,wang2010transportation,mun2008parsimonious,wang2010transportation}. \cite{sohn2006mobility} used coarse-grained GSM data to determine users' movements between staying in a place, walking and driving. By using boosted logistic regression in two phases, an accuracy of 80.85\% is reached for walking and driving. Although collecting GSM data is essentially a network-centric approach, data collection in this study mainly relied on limited number of lab members with a designed software for recording GSM records. Thus, the advantages of network-centric approaches are not fully in effect in this study. \cite{wang2010transportation} used coarse-grained unlabelled call detail records to infer transportation mode between pairs of defined origins and destinations. K-means algorithm was used in this study to detect mode of transportation. Low positioning accuracy, ping-pong handover effect, and privacy concerns have been mentioned in this study as some of the main problems of using GSM data. It should be noted that the GSM data is not readily available and needs cooperation from cellular network providers. \cite{mun2008parsimonious} coupled Wi-Fi and GSM data to reach a classification accuracy of 72.5\% for walking and driving in urban areas. They used decision tree to differentiate between walking, driving, and dwelling. Features used for classification in this experiment include Wi-Fi signal strength variance, duration of the dominant Wi-Fi access point, number of cell IDs that device connects to and residence time in cell footprint. Similar to Sohn's study, data collection in this study was designed as a controlled procedure using limited number of participants actively involved in the experiment. Unlike GPS data, for which accuracy is significantly affected in high-rise urban environments, Wi-Fi data can be implemented even in indoor environments. \cite{krumm2004locadio} used Wi-Fi signal strengths and their variance as inputs to a Hidden Markov Model for smoothing transitions between the inferred states of still and moving in indoor environments. An accuracy of 87\% was achieved in this study. The indoor environments are more controlled and have lesser heterogeneity, compared to outdoor environments. This study also did not consider transportation modes other than walking. \cite{beaulieu2019} used Wi-Fi data collected on a 14 blocks pedestrianized street in Mont\'eal to develop the next location choice model. They developed a dynamic mixed logit model with agent effects to achieve a maximum prediction performance of 70\%.

Growing interest in deep neural networks has led researchers to investigate deep architectures for the purpose of mode detection. \cite{endo2016deep} developed a deep neural network with fully connected layers to extract high-level features. Image-based deep features were combined with manual
features, and used as the input for a traditional classifier. Despite successfully implementing a deep network, the study lacks motion features such as speed and acceleration.
In another study, \cite{wang2017detecting} defined point features as a time series of speed, headway change, time interval, and distance between the GPS points. These features were then combined with manual features, and fed to a deep neural network. Despite using high-level features and developing deep neural networks, the accuracies obtained using these algorithms are still lower than some studies using manual data. Recently, \cite{dabiri2018inferring} developed a Convolutional Neural Network (CNN) architecture with different types of layers that are fed with an input layer with kinematic characteristics. Using their proposed architecture, a total accuracy of 84.8\% was achieved in this study. CNN architecture is well suited for user-centric approaches where the location is regularly sampled in terms of time. However, such an architecture is not suited for network-centric approaches where the user is observed at fixed locations that may or may not overlap.

In spite of promising performances of using deep neural networks for mode detection, all these studies are concentrated on labelled data from user-centric approaches, which requires vast amounts of labelled data. Semi-supervised learning is one of the possible approaches to address this issue by utilizing unlabelled data in order to enhance hypotheses obtained from labelled data~\cite{zhu05survey}. In short, semi-supervised algorithms add unlabelled samples to the training data and the classifier is retrained on the new augmented training dataset.

The advantages of Wi-Fi networks over other sources in terms of their granularity and ubiquity, along with the promising performance gains of deep residual networks in recent years, made us explore the feasibility of using Wi-Fi signals as the only source of data to detect transportation mode. Moreover, to overcome the difficulty of labelling in network-based approached, we introduced a semi-supervised version of the model.

\section{Case Study}
\label{S:4}
\begin{figure}[h]
\centering
\includegraphics[scale=0.65]{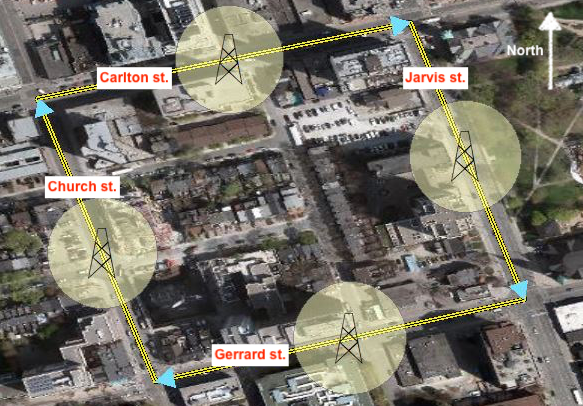}
\caption{Study area in downtown Toronto and OD\_Pods location}
\label{fig:odloc}
\end{figure}

We used \textit{URBANFlux system} that consists of a network of Wi-Fi detectors called OD\_Pods \cite{farooq2015ubiquitous} to record MAC addresses, signal strengths and times of connection for individual devices on four urban roads within the study area (Figure \ref{fig:odloc}). Coverage zone of each OD\_Pod can be approximated as a sphere with a radius of 50 meters. We collected the Wi-Fi signals from different smartphones on two separate days. The first wave of data collection was done on Wednesday, June 15, 2017, from 10 A.M. to 1 P.M. in a congested area in downtown Toronto. In order to increase the size of the data, and reduce human and experimental errors, the second round of data collection was done in the same area on Wednesday, August 22, 2018, from 10 A.M. to 1 P.M.

In both data collection rounds, parts of four downtown Toronto streets forming a loop were selected for installing OD\_Pods. This area was selected in order for the experiment to account for congested urban areas. As it is depicted in Figure \ref{fig:odloc}, the selected parts form a grid loop with a perimeter of 857 meters. Designated streets and locations of OD\_Pods were set so as to maintain the heterogeneity of the data. The designated loop includes a mix of separate bike lanes, side-walks, arterial, two-lane and one-lane streets. In addition, north edge of the loop is on the path of Toronto's 506 Carlton streetcars. Traffic signals on all four corners of the loop make the experiment more realistic and applicable to urban areas. The sensors are placed on mid-block locations so that no overlap occurs between coverage areas of OD\_Pods.

\subsection{Data Collection}
To collect labelled data, four volunteers were recruited, each going around the designated loop for 10 rounds. MAC address of participants' devices are recorded by \emph{URBANFlux system}. Modes are assigned to each participant in a way to resemble the actual share of different modes in downtown Toronto. To do so, the shares of walking, biking and driving trips are set to be approximately 25\%, 25\%, and 50\%. Data recorded from movements between each two OD\_Pods are considered as a trip with their respective mode. In addition, OD\_Pods automatically track MAC addresses of all Wi-Fi enabled devices moving in the area. MAC addresses different than those of the participants, which were recorded by at least two independent OD\_Pods during the experiment, were considered as the unlabelled trips. Table \ref{tab:Total} represents the total number of trips for each mode, along with unlabelled trips.

\begin{table}[h]
\centering
\label{tab:Tripsum}
\caption{Total number of trips collected for each mode}
\begin{tabular}{|l|c|}
\hline
    {\bf Mode} & {\bf Number of Trips} \\
\hline
\hline
{Walking} & 213  \\
{Biking} &        184     \\
{Driving} &        451     \\
{Unlabelled trips} &        1990     \\
\hline
\hline
Total & 2838\\
\hline
\end{tabular}
\label{tab:Total}
\end{table}

\subsection{Data Pre-processing}
Raw data extracted from OD\_Pods included three columns:
\begin{enumerate}
\item[a.] MAC address of a device
\item[b.] Signal strength
\item[c.] Time stamp
\end{enumerate}
All the data from different OD\_Pods are merged together. Having MAC addresses of participants, and non-participants detected by two independent OD\_Pods, connection data belonging to them are separated. Every movement between two OD\_Pods is considered as a trip observation, which includes data from origin OD\_Pod and destination OD\_Pod. Based on connection details from origin and destination OD\_Pods, 15 variables are derived as possible features for classification. These variables can be categorized to 3 groups:

\begin{enumerate}
\item[\textbf{a.}] \textbf{Time:} Relative travel speed and connection time variables are included in this group:

\textit{Relative Travel Speed} is represented by the normalized ratio of ``distance with no coverage zone between two involved OD\_Pods'' to its respective travel time, and is used as an indicator of travel speed. Speed-related variables have been the main features for classification in related literature \cite{poucin2018activity}. However, using speed solely does not guarantee satisfactory results. This can be explained by the fact that in congested areas, different transportation modes move at similar speeds. 

\textit{Connection time variables} represent variables related to the time that a device is communicating with the OD\_Pod in the coverage zone.
\item[\textbf{b.}] \textbf{Connection:} Number of communication messages exchanged with an OD\_Pod while a device is in its coverage zone. Intuitively, when a user spends more time in a coverage zone, for example, when walking in a coverage zone, the number of connections increases.
\item[\textbf{c.}] \textbf{Signal Strength:} Variance, first and second derivative of signal strengths during the time a device is communicating to an OD\_Pod. Intuitively, if a person passes through an OD\_Pod coverage are in short time, the fluctuations in signal strength are higher, which we are trying to observe in variance and derivatives of signal strength.
\end{enumerate}

\section{Methodology}
\label{S:3}
Here we develop a machine learning framework for mode detection that is specifically tailored to exploit the large, ubiquitous, low-cost, noisy, and partially labelled Wi-Fi data available in our case study. We use a semi-supervised residual net (ResNet) for developing very deep neural networks that are based on Multilayer Perceptron (MLP). Figure \ref{fig:Resarch} depicts the general working of the proposed framework. After extracting labelled and unlabelled data, a ResNet MLP classifier is trained using solely labelled data. The trained classifier is then used to predict the mode of transportation of a sample of unlabelled data. In the next step, predicted modes are used as the labels for the selected sample, known as \textit{Pseudo-labels}. Next, ResNet MLP classifier is re-trained on labelled data and pseudo-labelled data. The classifier is retrained on other portions of unlabelled data, and in the end, the accuracy of the method is evaluated on a labelled validation set.

In this study, the ResNet MLP architecture is developed and implemented as the classifier in the pseudo-labelling algorithm. In short, the classifier consists of an input layer, multiple building blocks, and an output layer. Each building block consists of two or three fully connected layers, with an identity shortcut connection that skips these layers.  

The proposed method in this research aims at addressing two challenging issues in related studies~\cite{he2016deep, lee2013pseudo}:
\begin{enumerate}
    \item[a.] Lowering the associated costs of data collection by incorporating a large amount of unlabelled data within the semi-supervised framework
    \item[b.] Benefiting from the large number of layers in a very deep network, without suffering from vanishing gradient problem, by implementing identity shortcut connection, derived from ResNet architecture
\end{enumerate}
In this section, we first describe ResNet and the way we developed our neural network using it. Pseudo-label, a simple, yet efficient semi-supervised learning method that we are applying to the core architecture is then discussed.

\subsection{ResNet-based MLP}
Deep Residual Network (ResNet) was originally introduced for image recognition and won the first place at the ILSVRC 2015 classification task \cite{he2016deep}. Although ResNet has been mainly implemented in convolutional neural networks, the idea of using \textit{identity shortcut connections} to skip one or more layers can be applied to other types of networks. In this study, we exploit this idea to develop a very deep multilayer perceptron.

The ResNet MLP developed for this study consists of three types of layers discussed below:
\begin{itemize}
    \item \textbf{Input Layer:} 15 features are extracted from raw labelled and unlabelled data. For each trip, values of features are normalized and inserted into the 15-node input layer.
    \item \textbf{Building Block:}
    A building block, or residual blocks, consists of a few fully connected layers and identity shortcuts. Finding adequate number of building blocks, and number of layers in each block, requires testing different networks and comparing the results.
    Feeding a block with input $x$, the output of the building block is \cite{he2016deep}:
    \begin{equation}
    \label{eq1}
        y= f(x,\{w_i\})+x
    \end{equation}
    in which $f$ is a shallow MLP, with two or three fully connected layers. For instance, in a two layer building block, $f$ can be written as:
    \begin{equation}
        f(x,\{w_1,w_2\})= W_2\sigma(W_1x) 
    \end{equation}
    Here $\sigma$ is the activation function, which is selected to be ReLU \cite{nair2010rectified}, similar to the original paper \cite{he2016deep}. 
    \item \textbf{Output Layer:} It consists of three nodes, for three modes of transportation investigated in our study. Similar to other layers in the network, output layer is fully connected. The outputs from previous layers are inserted into the last fully connected layer for classification. Softmax activation function is used to create probability distribution for three transportation modes.
\end{itemize}

By adding the output of shortcut connections to that of the stacked layers in equation \ref{eq1}, optimization of the network becomes easier, with no additional parameters or computational burden added \cite{he2016deep,he2016identity}. 

\subsection{Pseudo-Label}
Semi-supervised learning techniques are implemented to reduce the reliance on large amounts of expensive labelled data required in supervised learning algorithms. Unlabelled data are easy to collect in large volumes with lower costs compared to labelled datasets. In this study, a simple, yet efficient method for semi-supervised learning, called pseudo-label, is utilized \cite{lee2013pseudo}. 

In each iteration of the algorithm, pseudo-labels are defined as the classes with maximum predicted probability for each unlabelled sample.
The ResNet MLP network is trained on labelled and pseudo-labelled data in a supervised fashion. In every weight update, pseudo-labels are recalculated and are used in the loss function. The overall loss function of the learning task is written as \cite{lee2013pseudo}: 
\begin{equation}
    L=\frac{1}{n} \sum_{i=1}^{n} \sum_{m=1}^{M} L(y_i^m , f_i^m) + \alpha (t) \frac{1}{n^{'}} \sum_{i=1}^{n^{'}} \sum_{m=1}^{M} L(y^{'m}_i,f^{'m}_i)
\end{equation}
Here $n$ and $n'$ are the number of batches for labelled and unlabelled data, respectively; M is the number of classes i.e. modes of transportation investigated in our case; $y$ and $y'$ are labels and pseudo-labels respectively; $f$ and $f'$ are the network outputs of labelled and pseudo-labelled samples; and $\alpha(t)$ is the balancing coefficient \cite{lee2013pseudo}.

The performance of the algorithm varies for different sample rates. To find the best performance, various sample rates are tested, and the performances are compared.

Pseudo-labelling is a relatively simple semi-supervised algorithm based on self-learning scheme, that may perform poorly when the accuracy in predicting unlabelled samples are low. However, the simplicity and easy implementation of the algorithm, along with relatively high performance of the ResNet MLP classifier on labelled data, led us to add the algorithm to our ResNet MLP classifier. 

\begin{landscape}

\begin{center}

\begin{figure}
\vspace{35mm}

  \centering
  \includegraphics[scale=0.7]{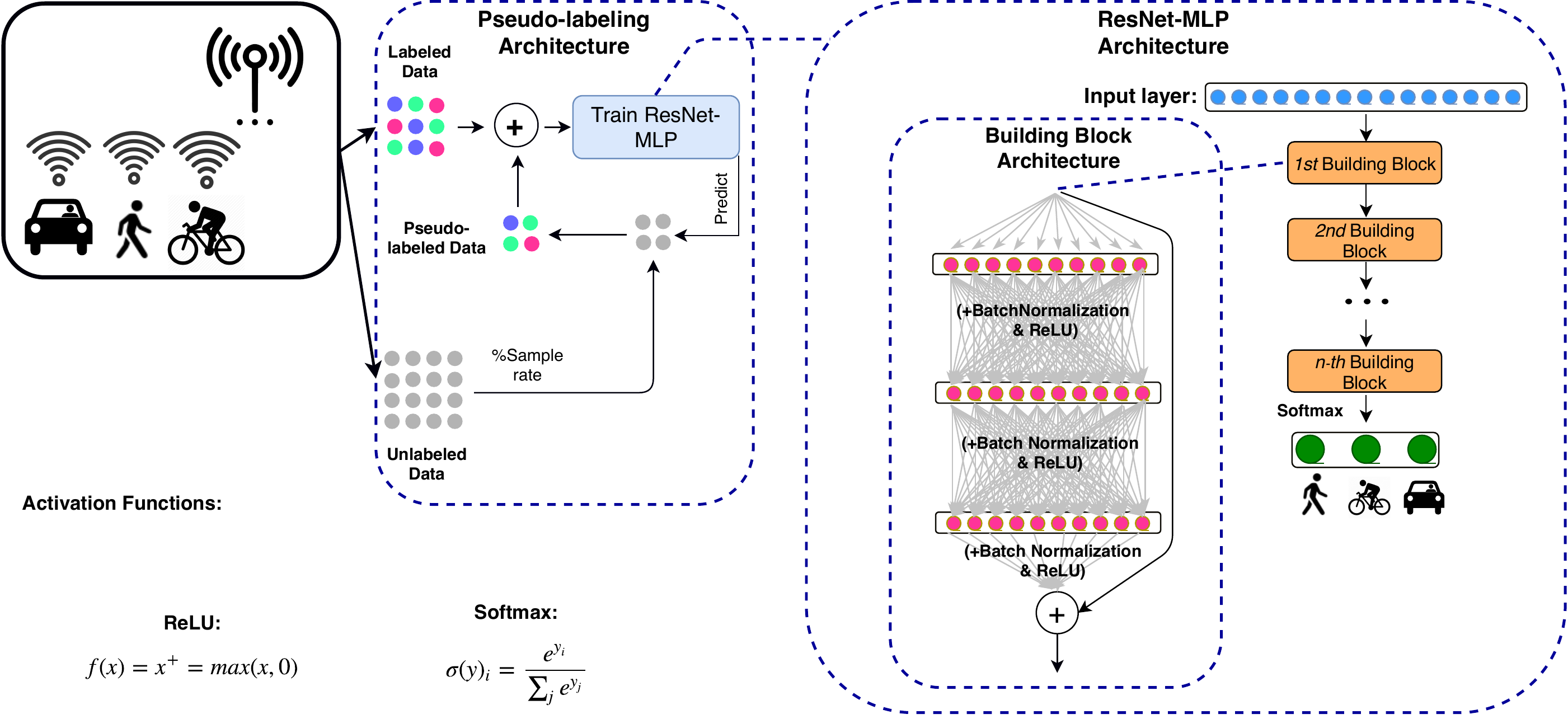}
  \caption{Conceptual framework}
  \label{fig:Resarch}
\end{figure}
\end{center}
\end{landscape}

\section{Results}
\label{S:5}
 For data preparation and feature extraction, R programming language is used. All the classification algorithms, including pseudo-labelling and ResNet MLP were coded in Python programming language, using Keras library \cite{chollet2015keras} and its implementation of TensorFlow with GPU support \cite{tensorflow2015-whitepaper}. After preparing the data for the analysis, network configuration is optimized based on ResNet34, which is one of the most commonly used architectures, initially introduced in~\cite{he2016deep}. In this step, whether to add batch normalization and dropout layers, and number of nodes (neurons) in each hidden layer is determined, and the best configurations are selected for the next stage. The best configurations selected in the previous step will then be implemented in different architectures, with different sample rates of pseudo-label algorithm to find the final model. 
 \subsection{Tuning ResNet Configuration}
 The ResNet core of the model can be configured in various ways. Going through all the possible conditions would require a timely and computationally expensive search. In this section, we try to cover a range of possible configurations based on dropout layers, batch normalization layers, and number of nodes (neurons) in hidden layers:
 \begin{itemize}
     \item \emph{Batch Normalization:} By performing normalization for each training mini-batch, batch normalization makes normalization a part of the model architecture. It also can be used as a regularizer to replace dropout in some cases \cite{ioffe2015batch}. 
     \item \emph{Dropout:} Aiming at reducing overfitting in the models, dropout's key idea is to randomly drop units and their connections from the neural network during training \cite{srivastava2014dropout}. 
     \item \emph{Number of Neurons (nodes):} Efficient number of nodes in each hidden layer should be carefully investigated, as low number of nodes can cause underfitting, and using too many nodes may result in overfitting. For the ease of calibration and to limit the number of possible architecture, we assume that all hidden layers are similar and have an equal number of nodes.
 \end{itemize}
As presented in Table \ref{tab:total}, 16 configurations can be generated based on the possible combinations of the three items. Each configuration is named using a letter \textit{a, b, c, d} representing number of nodes in hidden layers, and a regularization ID, all provided in parentheses in front of the items. For instance, Config-a10 refers to a configuration having 5 nodes in each hidden layer (a), with regularization ID \textit{10}, meaning that batch normalization is conducted before all layers (1) and no dropout layer is used (0).  All the configurations are implemented and tested in a supervised manner on ResNet34, which appears to be a good starting architecture, as it offers good performance in spite of less time complexity and model size comparing to other architectures in the original paper~\cite{he2016deep}. Performance of these configurations is compared after 500 number of epochs, with 20\%  of the data used as the validation set. For time-saving purposes, the accuracy of the models in this stage is not estimated using 10-fold cross validation. Batch size for all the tests is set equal to 32, and considering its successful performances in deep learning literature~\cite{kingma2014adam}, Adam method is used as the optimizer. Finally, best model will be selected for the next step, which will be the implementation of the core in a semi-supervised manner in different architectures. 

\begin{center}
\begin{table}[h]
\caption{Parameters of core ResNet architecture, values in () are for naming purposes}
\setlength\extrarowheight{2pt}
\centering
\begin{tabular}{|l|*{32}{p{\lena}|}}
\hline
\bf{Configuration Item}           & \multicolumn{32}{c|}{\bf{Explored Values}}\\
\hline\hline
Number of nodes in hidden layers & \mcii{5 (a)} & \mcii{10 (b)} & \mcii{15 (c)} & \mcii{20 (d)}  \\  
\hline
Batch Normalization & \mciii{No (0)} & \mciii{Yes (1)}  \\
\hline
Dropout Layer & \mciii{No (0)} & \mciii{Yes (1)}  \\
\hline
\end{tabular}
\label{tab:total}
\end{table}
\end{center}

\begin{figure}[t]
    \centering
    \includegraphics[scale=0.2]{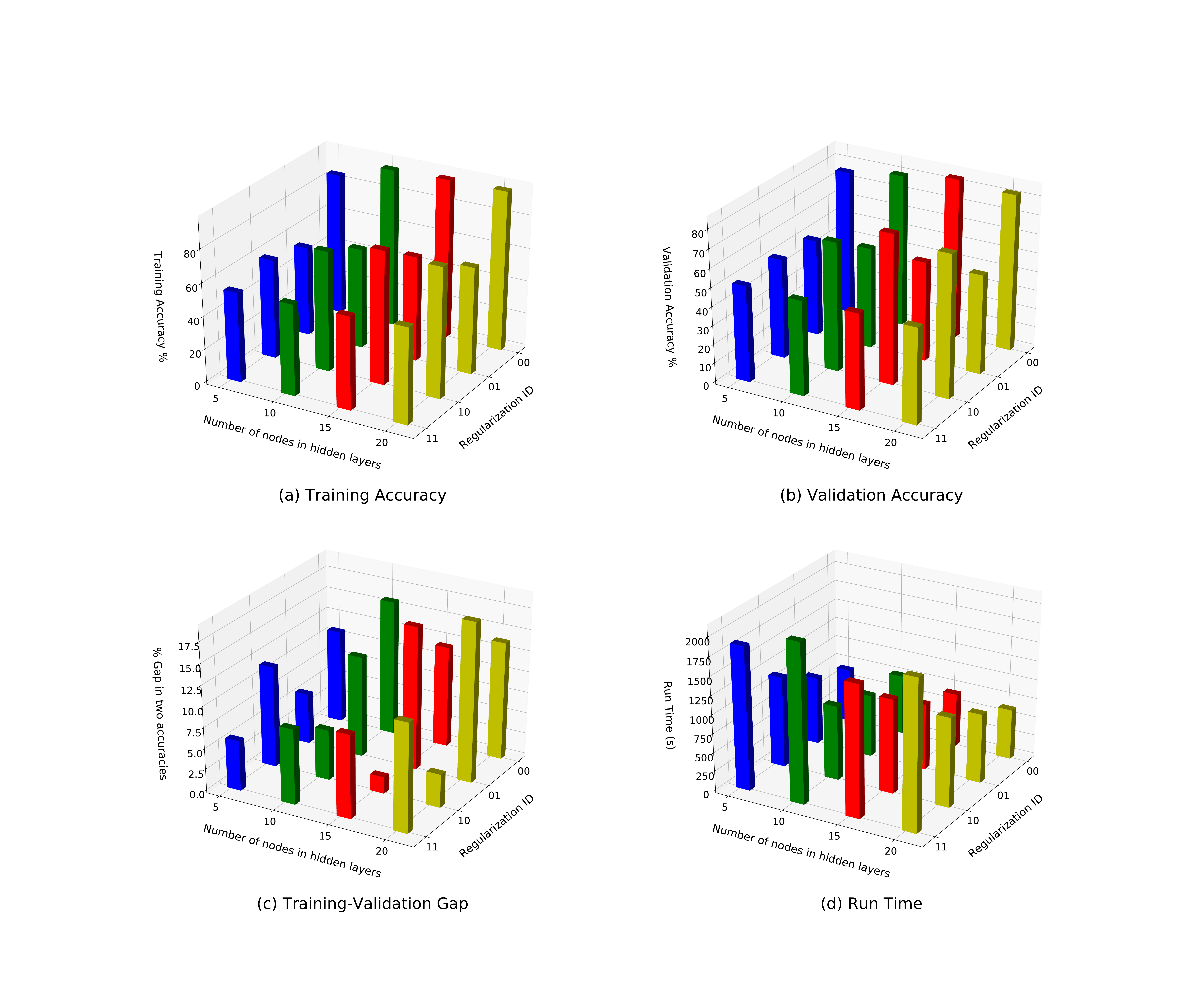}
    \caption{Comparing different ResNet34 Configurations }
    \label{fig:acc}
\end{figure}

Accuracy of ResNet34 on training set and validation set for all configurations, along with their respective run time and gap percentage between training and validation accuracy are provided in Table \ref{tab:acc}. Figure \ref{fig:acc} presents a comparative bar plots of the configurations. As shown in Figures \ref{fig:acc}a and \ref{fig:acc}b, all configurations having 20 nodes in their hidden layers, are performing worse on training and validation sets, than their counterpart configurations with 15 nodes in hidden layers. Thus, more nodes are not added to hidden layers. According to the results, top configurations based on accuracy of training and validation sets are: \textbf{c00}, \textbf{d00}, and \textbf{b00}. In all these three configurations, batch normalization and dropout layers are removed. The effect of removing regularization layers is clearly observable in gap \% of training and validation accuracy: despite high accuracy in training and validation sets, the gap between these two accuracies in three configurations are relatively high. The main purpose of batch normalization and dropout layer is to prevent overfitting over the training set. High accuracy on training data (over 95\% in three cases) and relatively large gap between training and validation accuracy is a sign of overfitting in these configurations. Configurations with regularization ID \textit{01}, in which dropout is used for regularization, have a maximum accuracy of 65\% on both training and validation sets. Considering high gap between validation and training accuracy, overfitting occurred in these configurations.  Adding batch normalization layers in configurations with regularization ID \textit{11}, helped to reduce overfitting slightly. However, the accuracies remained low.

\begin{table}[t]
    \caption{Different ResNet34 Configurations}
    \centering
    \small\addtolength{\tabcolsep}{-4pt}
    \begin{tabular}{|l|l|l|l|l|l|l|l|l|l|l|l|l|l|l|l|l|}
    \hline
       \textbf{Configuration}  & \textbf{a00} & \textbf{a01} & \textbf{a10} & \textbf{a11} & \textbf{b00} & \textbf{b01} & \textbf{b10} & \textbf{b11} & \textbf{c00} & \textbf{c01} & \textbf{c10} & \textbf{c11} &\textbf{d00}&\textbf{d01}&\textbf{d10}&\textbf{d11}  \\
       \hline\hline
    
         \textbf{Training accuracy (\%)}&85.7 &54.2 & 60.3  &54.7& 95.6 &60.7 &72.9& 55.7&96.6 &63.7 &81.1& 56.7& 96.9&65.1&79.3&  58.6 \\
    
         \textbf{Validation accuracy (\%)}&76.5 &51.2 & 52.9 & 51.2 & 80.6 &53.6&68.5 & 50.6&84.7 &52.9& 79.4& 51.2& 82.9&52.9&75.9& 51.2\\
        \textbf{\% of gap in two accuracies}& 11&6  & 12 &6 &16 &12 &6&9&12 &17 &2 &10  &14&19&4&13\\
         
    \hline
        \textbf{Run Time (s)}&698&889&1195&1895 & 783&820&985&2107 &703&875&1249&1752 & 662&915&1186&1999\\
        \hline
    
    \end{tabular}

    \label{tab:acc}
\end{table}
On the other hand, configurations with regularization ID \emph{10}, appear to have satisfying performance on both aspects: least gap between training and validation accuracy, and acceptable prediction accuracy on training and validation data.  Batch normalization in these configurations makes the networks not requiring dropout for regularization \cite{ioffe2015batch}.

\begin{figure}[t]
    \centering
    \includegraphics[scale=0.7]{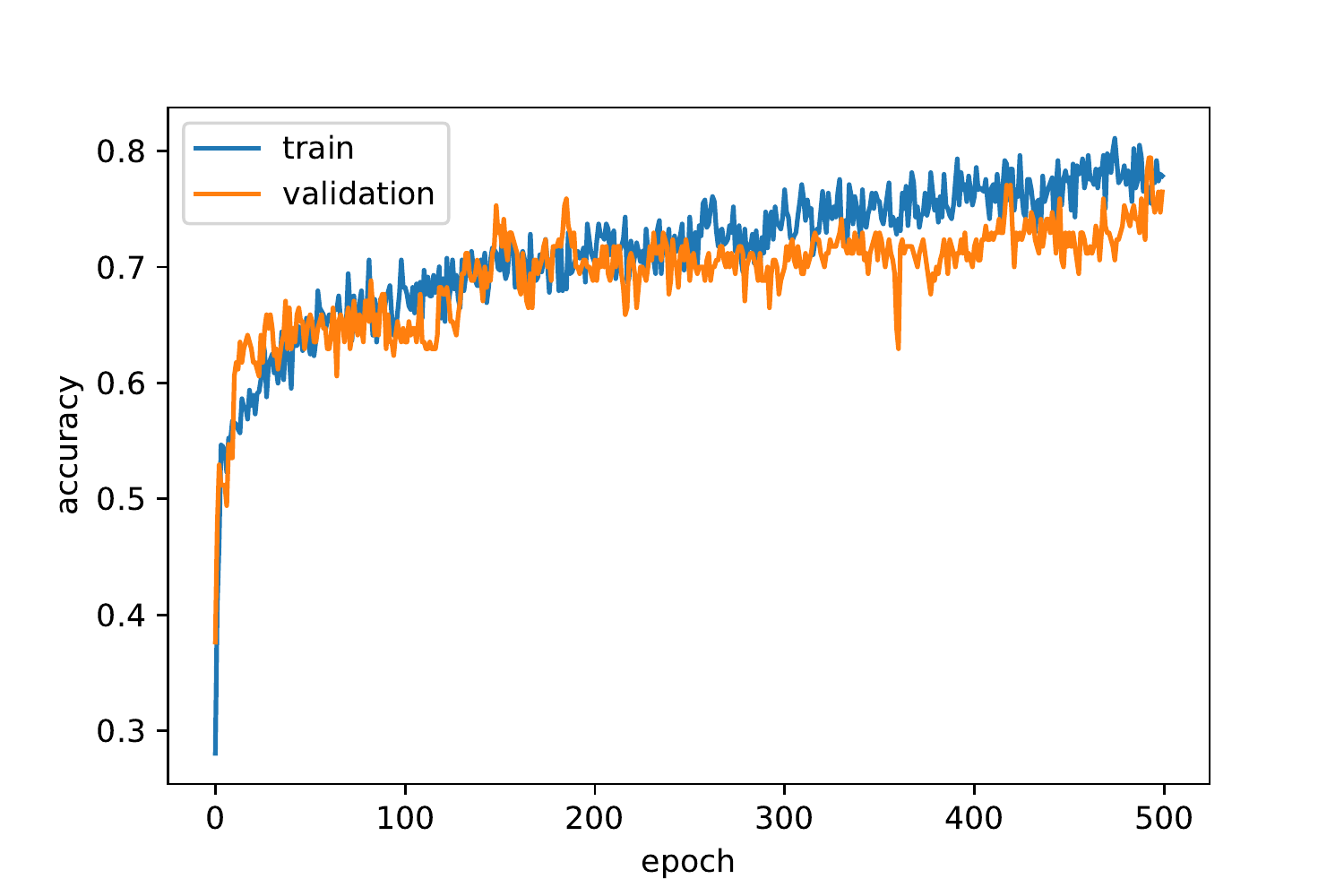}
    \caption{Training and validation performance of c10}
    \label{fig:epoch}
\end{figure}

Considering various aspects described above, \emph{c10} configuration is selected as the candidate in this step to be fed into our pseudo-label algorithm in the next stage.
To further investigate configurations, training run times are compared. As it is expected, adding dropout and batch normalization layers significantly increases the time it takes to run the code. However, for \emph{c10}, which is the configuration selected, it takes an affordable time of 1249 seconds to be trained by 500 epochs, on a 1.6GHz dual-core Intel Core i5 processor. Figure~\ref{fig:epoch} shows the trend of accuracy in training and validation set over 500 number of epochs for \emph{c10}.

 \subsection{Model Sensitivity Analysis}
 In the second stage for model calibration, configuration found in the previous section is fixed to find the best model architecture. In each architecture, number of layers within a building block is set to be equal to either 2 or 3 similar to \cite{he2016deep}. In addition, different number of building blocks within the ResNet MLP architecture are tested. Architectures are developed mainly based on their presence in~\cite{he2016deep}: For two-layer building block architectures, ResNet18 and ResNet34 and for three-layer building block architectures, ResNet50, ResNet101, and ResNet152 are tested. In addition, to explore other architectures, ResNet10 for two-layer building block networks, and ResNet74 and ResNet122 for three-layer building block networks are developed. The number in the model names represents the number of layer in that model. For instance, ResNet101 consists of 33 building blocks, each having three layers, plus an input and an output layer. On the pseudo-label part of the model, 6 sample rates are tested for each architecture: 0, 0.2, 0.4, 0.6, 0.8 and 1. Having a sample rate of 0 means that unlabelled data are not fed into the model, and pseudo-labelling part is ignored. By using a 0 sample rate, we will be able to better compare the performance of the classifier in a supervised and semi-supervised manner. On the other side, a sample rate of 1 means inserting all unlabeled data into the algorithm and pseudo-labeling them at once.  Model evaluation in this step is conducted using 10-fold cross-validation: dataset is randomly divided into 10 groups or folds. The first fold will be used as test set, and the other 9 folds will be used to train the classifier. This procedure is repeated 10 times, having a different fold as test set each time. Different model architectures are tested to find the optimal model with the best performance in 10-fold cross-validation. The number of epochs in training the model was set to be 1000 for all architectures. All the models are trained on a Core i7 4GHz CPU and a 16.0 GB memory.  
 
\begin{figure}[!h]
    \centering
    \begin{subfigure}{.476\textwidth}
        \centering
        \includegraphics[width=\linewidth]{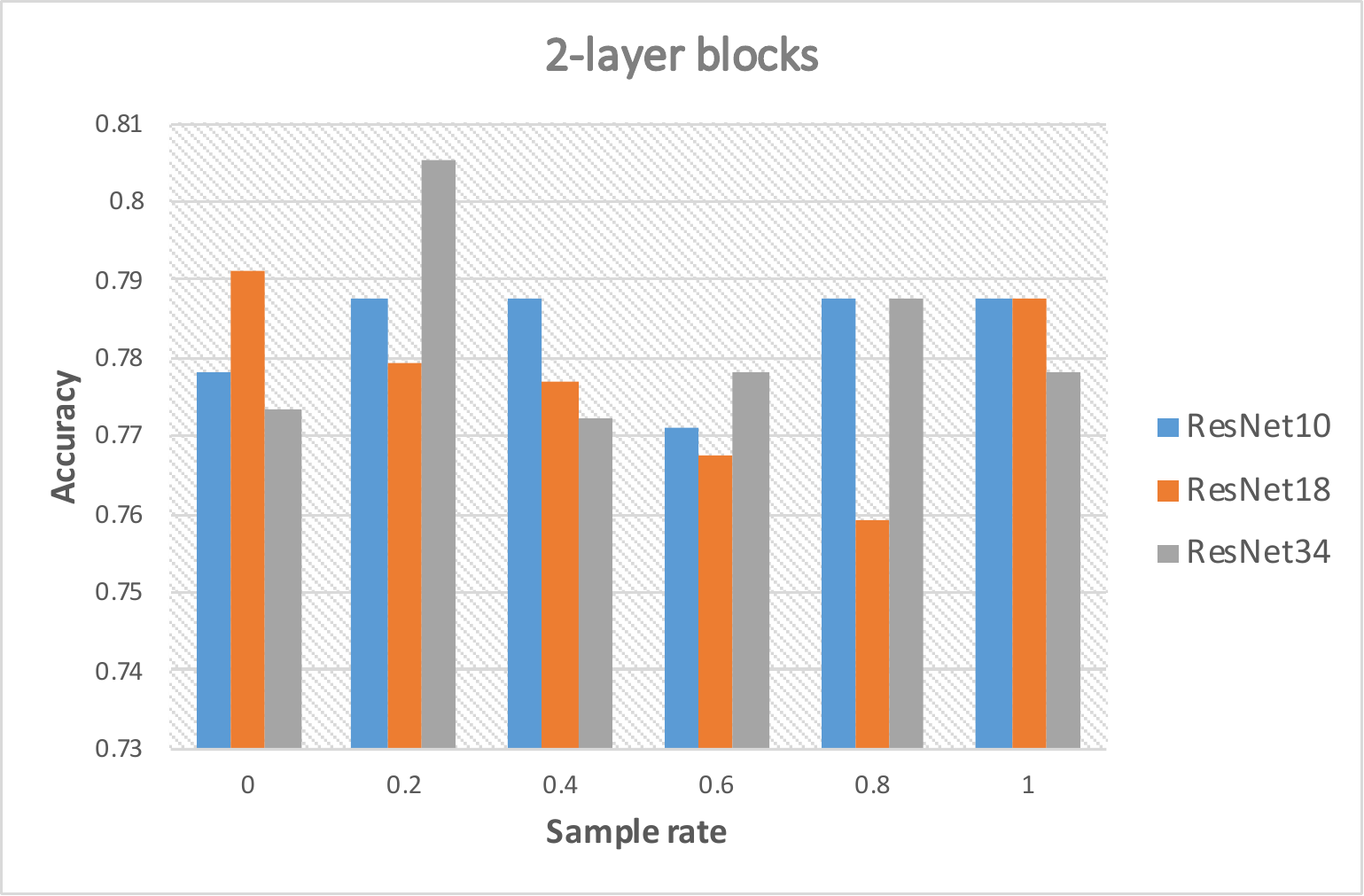}
        \caption{2-layer building blocks}
        \label{fig:results1}
    \end{subfigure}%
    \begin{subfigure}{.524\textwidth}
    \centering
    \includegraphics[width=0.98\linewidth]{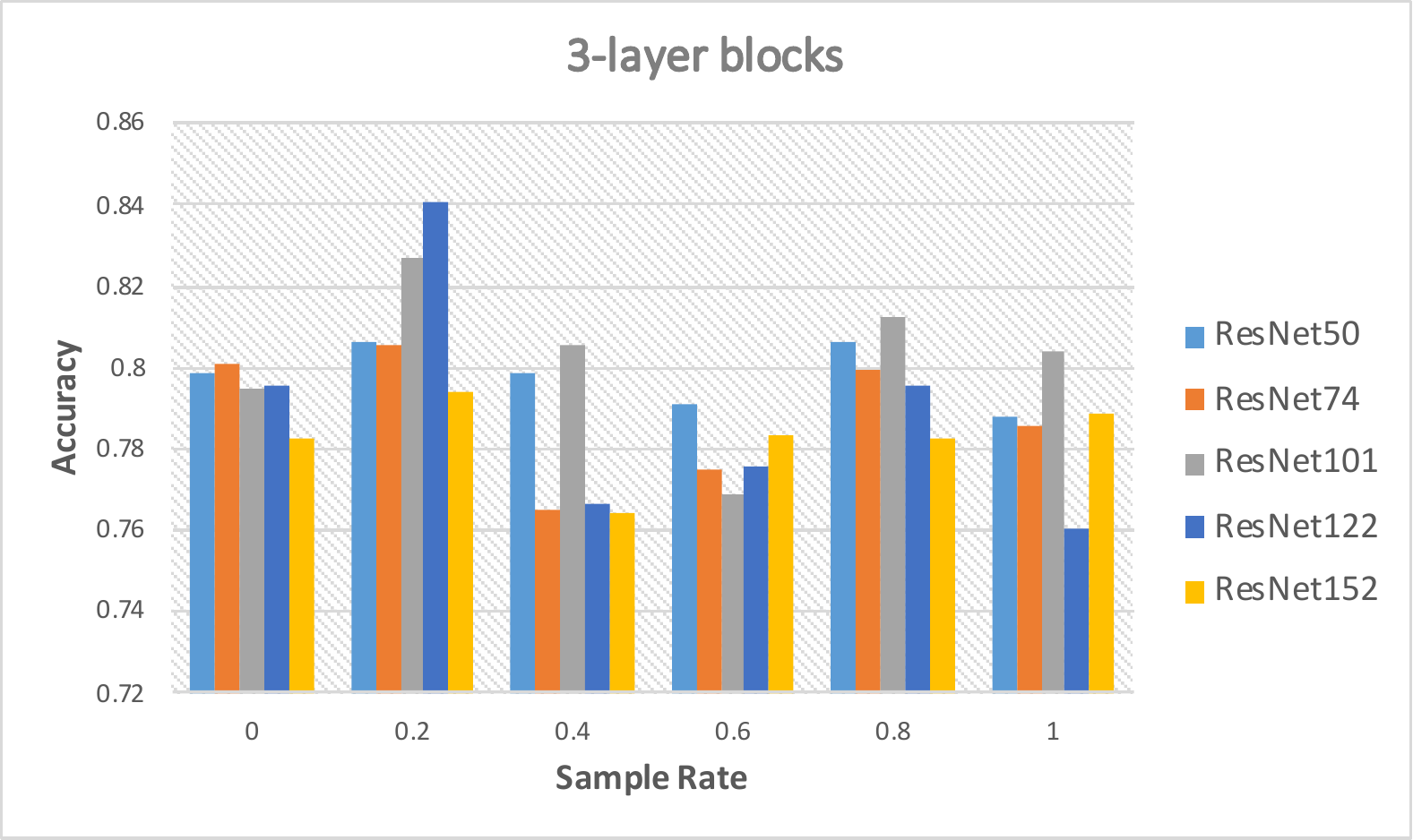}
    \caption{3-layer building blocks}
    \label{fig:results2}
    \end{subfigure}
    \caption{Model sensitivity analysis for number of building blocks and sampling rate}
\end{figure}

 Compared to plain non-residual architectures, all ResNet-based architectures showed significantly better performances than plain 34-layer MLP. A low accuracy of 56\% is achieved by using a deep non-residual network with 34 layers, which shows the power of adding identity shortcut connections to the network. Figure \ref{fig:results1} depicts the accuracy of different architectures for models with a two-layer building block. In figure \ref{fig:results2} three-layer blocks are compared.  As it can be inferred from the figures, performance of ResNet MLP architectures in a supervised manner, i.e. a sample rate of 0\%, appears not to be improving while increasing the number of hidden layers. ResNet18 for instance performs better than its deeper counterpart, ResNet32, for 2-layer building block models. Similarly, ResNet74 performs better than ResNet101 and ResNet122. However, adding the pseudo-label algorithm with a sample rate of 20\% results in a better performance of deeper networks. Specifically, for networks with more building blocks, improvement in performance using pseudo-labels appears to be more significant. For ResNet122 for instance, the accuracy increases by around 5\% with 20\% sample rate of pseudo-labels, whereas for ResNet10, the increase is less than 1\%, and for ResNet18, semi-supervised learning does not help to improve the performance. It can be concluded that having more building blocks, in general, tends to cause more overfitting, which is addressed by adding pseudo-labelled data to the model. 
 
 For sample rates greater than 20\%, the fluctuations in the accuracy increases and does not follow a predictable trend. Thus, sample rate of 20\% is selected as the optimum number for pseudo-labelling. On the other side, performance of the model in the deepest architecture, ResNet152 starts decreasing compared to ResNet122, which made us stop adding more hidden layers to the architecture. In conclusion, ResNet122 with a sample rate of 20\% is selected as the best model for classifying mode of transport using collected Wi-Fi data. Regarding the training time of the selected framework, a 10-fold cross-validation on the model takes around 4 hours to run, meaning an estimated average of 24 minutes per classifier.
 
\subsection{Final Model Accuracy Analysis}
The highest accuracy of the model, as it is depicted in Figure \ref{fig:results2} is achieved in ResNet122 with a sample rate of 20\%. With an 84.1\% accuracy in 10-fold cross-validation, the model outperforms all the other architectures, supervised and semi-supervised, residual and non-residual. In table \ref{tab:ConMat}, confusion matrix of this architecture is depicted. Recall and precision of each mode of transport in the confusion matrix are estimated based on a 20\% sample of labelled data, which was not used in the training procedure. Test set includes 170 rows of labelled data, including 44 walking, 40 biking and 86 driving trips. As it can be inferred from the table, the model successfully predicts all three modes of transportation with an accuracy of over 80\%. Among modes, driving has the most accurate recall and precision, despite the fact that both data collection experiments are conducted in congested urban areas, where the speed of vehicles are not significantly higher than bikes and pedestrians. Unlike commonly used mode classification methods which rely merely on features such as speed to distinguish trips, our model has benefited from higher level features with the help of very deep residual networks.

The lowest precision in the confusion matrix belongs to bikes. Biking and driving modes share many similar features, particularly in congested urban areas with signalized intersections. Thus, relatively low precision of biking can be explained. As it can be seen in Table \ref{tab:ConMat}, 8 driving trips that are incorrectly predicted as biking, play the main role in low accuracy of biking. These 8 trips consist less than 10\% of the driving trips. 
\begin{table}[!h]
\centering
\caption{Confusion Matrix of ResNet122 with 20\% sample rate} 
\small
\begin{tikzpicture}[
box/.style={draw,rectangle,minimum size=1.5cm,text width=1.5cm,align=center}]
\matrix (conmat) [row sep=.1cm,column sep=.1cm] {
\node (1) [box,
    label=left:\bf Walking,
    label=above:\bf Walking,
    ] {36};
&
\node (2) [box,
    label=above:\textbf{Biking},
    ] {3};
&
\node (3) [box,
    label=above:\textbf{Driving},
    ] {5};
&
\node(4) [label=above:\textbf{Total}]{44};
&
\node(5) [label=above:\textbf{Recall\%}]{81.8};
\\
\node (6) [box,
    label=left:\bf Biking,
    ] {2};
&
\node (7) [box,
    ] {33};
&
\node (8) [box,
    ] {5};
    &
\node(9) {40};
 &
\node(10) {82.5};
\\
\node (11) [box,
    label=left:\bf Driving,
    ] {4};
&
\node (12) [box,
    ] {8};
&
\node () [box,
    ] {74};
&
\node(14) {86};
 &
\node(15) {86.0};
\\
\node (16) [
    label=left:\textbf{Total},
    ] {42};
&
\node (17) {44};
&
\node (18)  {84};
&
\node (9)  {170};
\\
\node (10) [
    label=left:\textbf{Precision},
    ] {85.7};
&
\node (8) {75.0};
&
\node (9)  {88.1};
\\
};
\node [rotate=90,left=.1cm of conmat,anchor=center,text width=1.5cm,align=center] {\textbf{Actual}};
\node [above=.1cm of conmat,align=center,anchor=center] {\textbf{Prediction}};
\end{tikzpicture}
\label{tab:ConMat}
\end{table}
Due to a higher volume of vehicles in the area, our dataset consist of approximately 50\% driving trips. The higher number of driving trips in the data can result in the model to develop a tendency to be fitted better for this mode. This tendency led our framework to predict 5 walking trips (11\% of all walking trips) as driving. For a similar reason, 5 bike trips (12.5\% of all bike trips) are incorrectly predicted as driving. Despite all the above-mentioned inaccuracies, the model has been capable of predicting the correct mode of transportation with a satisfying accuracy of 84.1\% for all the trips based on 10-fold cross-validation results.

\section{Conclusion}
\label{S:6}
In this paper, a novel framework to infer transportation mode based on Wi-Fi data is introduced. By implementing \emph{URBANFlux} technology introduced in \cite{farooq2015ubiquitous}, Wi-Fi communication records in a congested urban area in downtown Toronto were collected on two separate days, as labelled data from participants and unlabelled data from non-participants. After data preparation processes and extracting fifteen features based on time and speed, signal strength and number of connections, a ResNet Multilayer Perceptron Network was developed and implemented as the core of a Pseudo-label semi-supervised learning algorithm. The proposed framework enables us to exploit the ample amount of low-cost unlabelled data, which makes our framework's performance adaptable to real life scenarios. In addition, by implementing a ResNet-based architecture, we could successfully train networks with a large number of hidden layer, which helps us benefit from high-level features in our classification task. Moreover, ResNet architecture helped us utilize noisy Wi-Fi data, with highly correlated features, which can be extracted with low costs. By adding semi-supervised learning to our framework, we tried to tackle the problems and costs of data collection. Calibrating this framework, we reached a total accuracy of 84.1\% for mode detection. 

Unlike other related works on the topic, in this study Wi-Fi data are not coupled with other sources of data to detect transportation mode in urban areas. This makes our proposed approach more cost-effective and easier to implement, with no interventions from users or processes like multi-source data fusion required. Another key differentiating feature in this study is the implementation of Wi-Fi-based mode detection on actual urban roads with real traffic at a reasonably large scale. Bluetooth and Wi-Fi signal sensors are already implemented in a number of cities, including Toronto. Labelling data collected from anonymous network users using these sensors can be extremely difficult and expensive, if not impossible. By implementing our framework, these data can be used in their unlabelled form to re-train our classifiers, and subsequently, improve their accuracy. Another advantage of our framework is its capability of being implemented in urban areas with Bluetooth or Wi-Fi sensors already incorporated, with virtually no additional cost for sensors and infrastructures.

With regard to the high penetration rate of smartphones in recent years, this method can be used within \emph{URBANFlux system} by city decision makers, operators, and planners to have a better understanding of users travel habits and their trends over time.  Transportation mode detection can also be useful in urban ubiquitous sensing, as it gives insight into energy consumption, pollution tracking and prediction and burned calorie estimation.

Our study is not without limitations which can be addressed in future studies. It can be extended by considering other different modes of transportation, i.e. subway, streetcars, and buses. This can be done by incorporating real-time data from transit schedules, which are available for most public transits in large cities. Moreover, underlying signal timing or network structure data can be added to further improve our knowledge on network users' modes of transportation. Real-time inference of transportation mode can also be investigated in future studies, adding unlabelled data from network users to update the framework continuously.  

\bibliographystyle{abbrv}
\bibliography{references.bib}

\begin{thebibliography}{10}

\bibitem{tensorflow2015-whitepaper}
M.~Abadi, A.~Agarwal, P.~Barham, E.~Brevdo, Z.~Chen, C.~Citro, G.~S. Corrado,
  A.~Davis, J.~Dean, M.~Devin, S.~Ghemawat, I.~Goodfellow, A.~Harp, G.~Irving,
  M.~Isard, Y.~Jia, R.~Jozefowicz, L.~Kaiser, M.~Kudlur, J.~Levenberg,
  D.~Man\'{e}, R.~Monga, S.~Moore, D.~Murray, C.~Olah, M.~Schuster, J.~Shlens,
  B.~Steiner, I.~Sutskever, K.~Talwar, P.~Tucker, V.~Vanhoucke, V.~Vasudevan,
  F.~Vi\'{e}gas, O.~Vinyals, P.~Warden, M.~Wattenberg, M.~Wicke, Y.~Yu, and
  X.~Zheng.
\newblock {TensorFlow}: Large-scale machine learning on heterogeneous systems,
  2015.
\newblock Software available from tensorflow.org.

\bibitem{beaulieu2019}
A.~Beaulieu and B.~Farooq.
\newblock A dynamic mixed logit model with agent effect for pedestrian next
  location choice using ubiquitous wi-fi network data.
\newblock {\em International Journal of Transportation Science and Technology},
  2019.

\bibitem{blum1998combining}
A.~Blum and T.~Mitchell.
\newblock Combining labeled and unlabeled data with co-training.
\newblock In {\em Proceedings of the eleventh annual conference on
  Computational learning theory}, pages 92--100. ACM, 1998.

\bibitem{chen2010evaluating}
C.~Chen, H.~Gong, C.~Lawson, and E.~Bialostozky.
\newblock Evaluating the feasibility of a passive travel survey collection in a
  complex urban environment: Lessons learned from the new york city case study.
\newblock {\em Transportation Research Part A: Policy and Practice},
  44(10):830--840, 2010.

\bibitem{chollet2015keras}
F.~Chollet et~al.
\newblock Keras.
\newblock \url{https://keras.io}, 2015.

\bibitem{dabiri2018inferring}
S.~Dabiri and K.~Heaslip.
\newblock Inferring transportation modes from gps trajectories using a
  convolutional neural network.
\newblock {\em Transportation research part C: emerging technologies},
  86:360--371, 2018.

\bibitem{endo2016deep}
Y.~Endo, H.~Toda, K.~Nishida, and A.~Kawanobe.
\newblock Deep feature extraction from trajectories for transportation mode
  estimation.
\newblock In {\em Pacific-Asia Conference on Knowledge Discovery and Data
  Mining}, pages 54--66. Springer, 2016.

\bibitem{farooq2015ubiquitous}
B.~Farooq, A.~Beaulieu, M.~Ragab, and V.~D. Ba.
\newblock Ubiquitous monitoring of pedestrian dynamics: Exploring wireless ad
  hoc network of multi-sensor technologies.
\newblock In {\em SENSORS, 2015 IEEE}, pages 1--4. IEEE, 2015.

\bibitem{gong2012gps}
H.~Gong, C.~Chen, E.~Bialostozky, and C.~T. Lawson.
\newblock A gps/gis method for travel mode detection in new york city.
\newblock {\em Computers, Environment and Urban Systems}, 36(2):131--139, 2012.

\bibitem{he2016deep}
K.~He, X.~Zhang, S.~Ren, and J.~Sun.
\newblock Deep residual learning for image recognition.
\newblock In {\em Proceedings of the IEEE conference on computer vision and
  pattern recognition}, pages 770--778, 2016.

\bibitem{he2016identity}
K.~He, X.~Zhang, S.~Ren, and J.~Sun.
\newblock Identity mappings in deep residual networks.
\newblock In {\em European conference on computer vision}, pages 630--645.
  Springer, 2016.

\bibitem{ioffe2015batch}
S.~Ioffe and C.~Szegedy.
\newblock Batch normalization: Accelerating deep network training by reducing
  internal covariate shift.
\newblock {\em arXiv preprint arXiv:1502.03167}, 2015.

\bibitem{kingma2014adam}
D.~P. Kingma and J.~Ba.
\newblock Adam: A method for stochastic optimization.
\newblock {\em arXiv preprint arXiv:1412.6980}, 2014.

\bibitem{krumm2004locadio}
J.~Krumm and E.~Horvitz.
\newblock Locadio: Inferring motion and location from wi-fi signal strengths.
\newblock In {\em mobiquitous}, pages 4--13, 2004.

\bibitem{lee2013pseudo}
D.-H. Lee.
\newblock Pseudo-label: The simple and efficient semi-supervised learning
  method for deep neural networks.
\newblock 2013.

\bibitem{mun2008parsimonious}
M.~Mun, D.~Estrin, J.~Burke, and M.~Hansen.
\newblock Parsimonious mobility classification using gsm and wifi traces.
\newblock In {\em Proceedings of the Fifth Workshop on Embedded Networked
  Sensors (HotEmNets)}, 2008.

\bibitem{murakami2004using}
E.~Murakami, D.~P. Wagner, and D.~M. Neumeister.
\newblock Using global positioning systems and personal digital assistants for
  personal travel surveys in the united states.
\newblock 2004.

\bibitem{nair2010rectified}
V.~Nair and G.~E. Hinton.
\newblock Rectified linear units improve restricted boltzmann machines.
\newblock In {\em Proceedings of the 27th international conference on machine
  learning (ICML-10)}, pages 807--814, 2010.

\bibitem{poucin2018activity}
G.~Poucin, B.~Farooq, and Z.~Patterson.
\newblock Activity patterns mining in wi-fi access point logs.
\newblock {\em Computers, Environment and Urban Systems}, 67:55--67, 2018.

\bibitem{reddy2008determining}
S.~Reddy, J.~Burke, D.~Estrin, M.~Hansen, and M.~Srivastava.
\newblock Determining transportation mode on mobile phones.
\newblock In {\em Wearable computers, 2008. ISWC 2008. 12th IEEE International
  symposium on}, pages 25--28. IEEE, 2008.

\bibitem{reed2014training}
S.~Reed, H.~Lee, D.~Anguelov, C.~Szegedy, D.~Erhan, and A.~Rabinovich.
\newblock Training deep neural networks on noisy labels with bootstrapping.
\newblock {\em arXiv preprint arXiv:1412.6596}, 2014.

\bibitem{rolnick2017deep}
D.~Rolnick, A.~Veit, S.~Belongie, and N.~Shavit.
\newblock Deep learning is robust to massive label noise.
\newblock {\em arXiv preprint arXiv:1705.10694}, 2017.

\bibitem{sohn2006mobility}
T.~Sohn, A.~Varshavsky, A.~LaMarca, M.~Chen, T.~Choudhury, I.~Smith,
  S.~Consolvo, J.~Hightower, W.~Griswold, and E.~De~Lara.
\newblock Mobility detection using everyday gsm traces.
\newblock {\em UbiComp 2006: Ubiquitous Computing}, pages 212--224, 2006.

\bibitem{srivastava2014dropout}
N.~Srivastava, G.~Hinton, A.~Krizhevsky, I.~Sutskever, and R.~Salakhutdinov.
\newblock Dropout: a simple way to prevent neural networks from overfitting.
\newblock {\em The Journal of Machine Learning Research}, 15(1):1929--1958,
  2014.

\bibitem{stenneth2011transportation}
L.~Stenneth, O.~Wolfson, P.~S. Yu, and B.~Xu.
\newblock Transportation mode detection using mobile phones and gis
  information.
\newblock In {\em Proceedings of the 19th ACM SIGSPATIAL International
  Conference on Advances in Geographic Information Systems}, pages 54--63. ACM,
  2011.

\bibitem{stopher2007household}
P.~R. Stopher and S.~P. Greaves.
\newblock Household travel surveys: Where are we going?
\newblock {\em Transportation Research Part A: Policy and Practice},
  41(5):367--381, 2007.

\bibitem{opendata}
T.~C.~o. Toronto.
\newblock Travel times - bluetooth, retrieved from
  https://www.toronto.ca/city-government/data-research-maps/open-data/,
  2014-2017.

\bibitem{wang2010transportation}
H.~Wang, F.~Calabrese, G.~Di~Lorenzo, and C.~Ratti.
\newblock Transportation mode inference from anonymized and aggregated mobile
  phone call detail records.
\newblock In {\em Intelligent Transportation Systems (ITSC), 2010 13th
  International IEEE Conference on}, pages 318--323. IEEE, 2010.

\bibitem{wang2017detecting}
H.~Wang, G.~Liu, J.~Duan, and L.~Zhang.
\newblock Detecting transportation modes using deep neural network.
\newblock {\em IEICE TRANSACTIONS on Information and Systems},
  100(5):1132--1135, 2017.

\bibitem{xiao2017identifying}
Z.~Xiao, Y.~Wang, K.~Fu, and F.~Wu.
\newblock Identifying different transportation modes from trajectory data using
  tree-based ensemble classifiers.
\newblock {\em ISPRS International Journal of Geo-Information}, 6(2):57, 2017.

\bibitem{zheng2008understanding}
Y.~Zheng, Q.~Li, Y.~Chen, X.~Xie, and W.-Y. Ma.
\newblock Understanding mobility based on gps data.
\newblock In {\em Proceedings of the 10th international conference on
  Ubiquitous computing}, pages 312--321. ACM, 2008.

\bibitem{zhu05survey}
X.~Zhu.
\newblock Semi-supervised learning literature survey.
\newblock Technical Report 1530, Computer Sciences, University of
  Wisconsin-Madison, 2005.

\end{thebibliography}
\end{document}